\documentclass[
sigconf = true, 
review = false, 
screen = true, 
anonymous = false, 
]{acmart}

\AtBeginDocument{%
  \providecommand\BibTeX{{%
    \normalfont B\kern-0.5em{\scshape i\kern-0.25em b}\kern-0.8em\TeX}}}

\copyrightyear{2024}
\acmYear{2024}
\setcopyright{rightsretained}
\acmConference[GECCO '24]{Genetic and Evolutionary Computation
Conference}{July 14--18, 2024}{Melbourne, VIC, Australia}
\acmBooktitle{Genetic and Evolutionary Computation Conference (GECCO '24),
July 14--18, 2024, Melbourne, VIC, Australia}
\acmDOI{10.1145/3638529.3654176}
\acmISBN{979-8-4007-0494-9/24/07}
\acmPrice{15.00}

\usepackage{algorithm}
\usepackage{algorithmic}

\usepackage{subcaption}

\usepackage{xspace}

\newcommand{\gs}{GSEMO\xspace}
\newcommand{\sg}{SW-GSEMO\xspace}
\newcommand{\asg}{ASW-GSEMO\xspace}

\DeclareMathOperator*{\argmax}{argmax}

\begin{document}

\title{
Sampling-based Pareto Optimization for Chance-constrained Monotone Submodular Problems}

\author{Xiankun Yan}
\orcid{0000-0002-2309-8034}
\affiliation{%
   \institution{Optimisation and Logistics,\\
   School of Computer and Mathematical Sciences,\\
   The University of Adelaide
   \city{Adelaide}
   \country{Australia}}
}

\author{Aneta Neumann}
\orcid{0000-0002-0036-4782}
\affiliation{%
  \institution{Optimisation and Logistics,\\
   School of Computer and Mathematical Sciences,\\
   The University of Adelaide
   \city{Adelaide}
   \country{Australia}}
  }

\author{Frank Neumann}
\orcid{0000-0002-2721-3618}
\affiliation{%
   \institution{Optimisation and Logistics,\\
   School of Computer and Mathematical Sciences,\\
   The University of Adelaide
   \city{Adelaide}
   \country{Australia}}
}

\renewcommand{\shortauthors}{Xiankun Yan, Aneta Neumann and Frank Neumann}

\begin{abstract}
 Recently surrogate functions based on the tail inequalities were developed to evaluate the chance constraints in the context of evolutionary computation
  and several Pareto optimization algorithms using these surrogates were successfully applied in optimizing chance-constrained monotone submodular problems.
  However, the difference in performance between algorithms using the surrogates and those employing the direct sampling-based evaluation remains unclear.
  Within the paper, a sampling-based method is proposed to directly evaluate the chance constraint. 
  Furthermore, to address the problems with more challenging settings,
  an enhanced GSEMO algorithm integrated with an adaptive sliding window, called ASW-GSEMO, is introduced. 
  In the experiments, the \asg employing the sampling-based approach is tested on the chance-constrained version of the maximum coverage problem with different settings.
  Its results are compared with those from other algorithms using different surrogate functions. 
  The experimental findings indicate that the \asg with the sampling-based evaluation approach outperforms other algorithms,
  highlighting that the performances of algorithms using different evaluation methods are comparable. 
  Additionally, the behaviors of \asg are visualized to explain the distinctions between it and the algorithms utilizing the surrogate functions.   
  
\end{abstract}

\ccsdesc[500]{Mathematics of computing~Stochastic control and optimization}

\keywords{Chance-constrained submodular problem, fast Pareto optimization, sampling-based evaluation, adaptive sliding-selection}


\maketitle

\section{Introduction}
In real-world optimization problems, diminishing returns often affect the profit of a solution when adding more items to the solution. 
Such problems can be generalized by the submodular functions as highlighted in previous research~\cite{nemhauser1978best,nemhauser1978analysis,krause2014submodular}.
In the past, the optimization of monotone submodular functions under deterministic knapsack constraint has been widely studied~\cite{nemhauser1978analysis,khuller1999budgeted,doskovc2020non,neumann2021diversifying}. 
The aim of such problems is to find a subset of elements that maximizes the function value without the total weight of the subset exceeding a bound given in the knapsack. 
Recently, people gradually paid more attention to the stochastic version of the optimization of monotone submodular functions since the stochastic scenarios in real-world applications are unavoidable.
Consequently, it is crucial not only to consider the existing constraints but also to minimize the effects of stochastic items in the stochastic optimization problems. 

A useful technique, \emph{chance constraint}, is frequently utilized to address the effects of uncertainty in the problem involving stochastic items~\cite{nemirovski2006scenario,xie2019evolutionary,assimi2020evolutionary,DBLP:conf/gecco/XieN020,DBLP:conf/gecco/XieN0S21,DBLP:conf/ppsn/NeumannXN22,shi2022runtime,yan2023optimizing,E23OMOEA,U3OEADCHKP,TCHTTP,MOEAwSWS,ERDCCCMCP}. 
This approach allows for a solution's actual weight to exceed the given bound but with a very small probability. 
Previous studies~\cite{doerr2020optimization,neumann2021diversifying,neumann2020optimising} have successfully applied the chance constraint to the optimization of monotone submodular functions under the uniform distribution setting.
Furthermore, to evaluate the chance constraint, surrogate functions were established based on One-sided Chevbyshev's inequality and Chernoff bound,
These surrogate functions facilitate the translation of the chance constraint into their corresponding deterministic equivalents for a given confidence level.
Additionally, the performance of some greedy algorithms and a global simple evolutionary multi-objective algorithm (\gs) were analyzed and tested on the chance-constrained submodular problem. 
From their theoretical analysis, it is shown that within a certain runtime,
all algorithms using the surrogate functions can guarantee a solution with a good approximation relative to the optimal solution obtained in the corresponding deterministic setting.  
However, the previous work did not address the gap between the algorithm using the surrogate functions and those employing a sampling-based method to directly evaluate the chance constraint. 
Furthermore, the performance of algorithms in more complex settings has yet to be explored. 

The sampling-based approach proves valuable for evaluating the chance constraint~\cite{luedtke2008sample,ahmed2008solving,pena2020solving}.
In the paper, we introduce a novel sampling-based method for optimizing chance-constrained submodular problems.
This method involves sampling the actual weights of a solution multiple times and then sorting these weights to identify the highest sampled weight that complies with the specified probability threshold.
If this maximum sampled weight remains below the bound given in the chance constraint, the solution is regarded as feasible; otherwise, it is considered infeasible.
In addition, we propose an enhanced \gs integrated with an adaptive window inspired by the sliding window \gs (\sg) in \cite{Neumann2023fast}.
This advanced algorithm is designated as \asg, to tackle problems under challenging settings. 
In \cite{Neumann2023fast}, Neumann and Witt explain how the defined window in the \sg helps the algorithm in selecting some potential individuals for the parent selection phase. 
It is important to note that the size of the window is fixed as 1, and the algorithm is tested in deterministic settings with smaller weights.  
However, our experiments reveal that the \sg struggles to capture any individuals within the window when dealing with larger weights. 
To address this issue, the adaptive window is proposed.
In the adaptive window, the lower bound of the window is still determined based on the given bound and the ratio between the current time and total time.
However, the window size is allowed to increase by 1 step whenever no individual is present within the window.
Simultaneously, inspired by the theoretical analysis of the sliding-selection presented in \cite{Neumann2023fast}, 
the adaptive window aims to capture only one individual during optimization.
Therefore, the window size is maintained when only one individual is within the window.
However, if more than one individual is available for selection, the window size is reduced by 1 in the subsequent iteration.
Moreover, following the experimental methodology in~\cite{neumann2020optimising}, 
we test the performances of the \asg using the sampling-base approach on the maximum coverage problems.
The problems are based on the graphs under two different settings,
including independent and identically distributed (IID) weights and uniform weights with the same dispersion. 
The experiments involve visualizing the behavior of \asg and comparing its performance against the \gs and \sg across various instances and sampling sizes.
Additionally, we explore the gap between results obtained from the sampling-based approach and those derived from different surrogate methods.
The empirical findings indicate that the \asg outperforms other algorithms when the sampling-based approach is employed, 
and the quality of solutions obtained through this approach is comparable with those using surrogate methods.

The structure of this paper is organized in the following.
Section~\ref{sec:pds} outlines the definition of the chance-constrained submodular problem and the settings investigated for the experiments. 
Section~\ref{sec:pre_app} discusses the previously established surrogate evaluation methods and the algorithms as the baseline applied to the problem under various settings. 
Section~\ref{sec:fspo} introduced the proposed sampling-based evaluation method and the \asg. 
Section~\ref{sec:exp} investigates the performance analysis of the different algorithms using various evaluation methods and includes a visualization of how the \asg functions during optimization.
Finally, Section~\ref{sec:conclu} summarizes our findings and discusses future work.

\section{Problem Definition and Settings}
\label{sec:pds}
Given a ground set $V = \{v_1,...,v_n\}$, the optimzation of a monotone submodular function $f: 2^V\to \mathbb{R}_{\geq 0}$ is considered.
A function is defined as monotone iff for every $S\subseteq T\subseteq V$, $f(S)\leq f(T)$ holds.
Besides, a function $f$ is submodular iff for every $S\subseteq T\subseteq V$ and $v\notin T$, 
$f(S\cup \{v\})-f(S) \geq f(T\cup \{v\})-f(T)$ is satisfied. 
Within the paper, the optimization of a monotone submodular function $f$ is considered
subjected to the chance constraint where each element $v_i\in V$ has a stochastic weight $W(v_i)$. 
The chance-constrained  optimization problem can be formulated as
\begin{align*}
    Maximize & \quad f(S) \\
    S.t. & \quad Pr[W(S)>B]\leq \alpha,
\end{align*}
where $W(S)$ is the weight of the subset $S$ (i.e., $W(S) = \sum_{v_i\in S} W(v_i)$) and 
$B$ is the deterministic bound. 
The parameter $\alpha$ is a small and acceptable probability violating the bound $B$.

For the settings, they are both considered the actual weight of the element sampled from a uniform distribution with an expected weight and dispersion. 
There are two different cases for the expected weight $E_W(v_i)$ and dispersion $\delta(v_i)$ investigated in the paper:
\begin{enumerate}
    \item \emph{Uniform IID Weights} the stochastic weight $W(v_i)$ has the same expected weight and same dispersion, i.e., $E_W(v_i)=a$, $\delta(v_i) = d$, and $W(v_i) \in [a-d,a+d$] with $0<d\leq a$. 
    \item \emph{Uniform Weights with the Same Dispersion} the elements' expected weights are different but their dispersion is the same, i.e, $E_W(v_i)=a_i$, $\delta(v_i) = d$, and $W(v_i) \in [a_i-d,a_i+d$] with $0<d\leq a_i$.
\end{enumerate}

Since the evolutionary algorithm is considered to solve the problem,
we encode an element set $S$ as a decision vector $x = x_1 x_2...x_n$ with length $n$, 
where $x_i=1$ means that the element $v_i \in V$ is picked up into the solution $S$. 
Given that all settings are based on a uniform distribution, the expected weight and variance of the solution are calculated as follows:
$E[W(X)] = \sum_{i=0}^n E_W(v_i)x_i,$
and
$Var[W(X)] = \sum_{i=0}^n \delta(v_i)^2x_i/3.$

\section{Related work}
\label{sec:pre_app}
As described in previous work~\cite{neumann2020optimising},
the chance constraint is evaluated using the surrogate functions based on \emph{One-sided Chebyshev's inequality} and \emph{Chernoff bound}.
The surrogate weight calculated by these different surrogate functions can be respectively expressed as
$$W_{cheb}(X) = E[W(X)] + \sqrt{\frac{(1-\alpha)Var[W(X)]}{\alpha}},$$
and 
$$W_{chen}(X) = E[W(X)] + \sqrt{3d|X|\ln{(1/\alpha)}},$$
where $|X|$ is the number of elements picked in the solution. 
Furthermore, multi-objective evolutionary algorithms (MOEAs) are studied to optimize the given problem in~\cite{neumann2020optimising}.
In the algorithm, each solution is considered a two two-dimensional search point in the objective space.
The two-dimensional fitness function of the solution $X$ is expressed as
\begin{equation}
    g_1(X) = \left\{
    \begin{array}{ccl}
        f(X) & & {g_2(X)\leq B}  \\
        -1 & & {g_2(X)>B}
    \end{array}\right.
\end{equation}

\begin{equation}
    g_2(X) =
        W_{sg}(X),
\end{equation}
where $f(X)$ is the submodular function value of $X$, $W_{sg}(X)$ is the surrogate weight of $X$.
Let $Y \in \{0,1\}^n$ be another solution in the search space.
Solution $X$ (weakly) dominates $Y$ (denoted as $X\succeq Y$) iff $g_1(X) \geq g_1(Y)$ and $g_2(X) \leq g_2(Y)$.
Comparatively, an infeasible solution is strongly dominated by a feasible one due to the objective function $g_1$.
Additionally, the objective function $g_2$ directs the solutions towards the feasible search space.

In~\cite{neumann2020optimising}, the \gs (see in Algorithm~\ref{alg:gsemo}) is employed to address the chance-constrained problem.
In our study, we consider this algorithm with the defined two-objective fitness function,
as the baseline for subsequent experiments.
The \gs initiates with an empty set and progressively includes items into this set by utilizing the standard bit mutation operator throughout the optimization process.
During mutation, a new solution is generated by independently flipping each bit of the existing solution with a probability of $1/n$.
If the newly generated offspring is dominated by any individual in the current population, it is subsequently rejected.
Additionally, the \gs maintains a set of solutions,
each of which is not dominated by any solution produced at each iteration during the optimization.

The \sg (given in Algorithm~\ref{alg:sw_gsemo}) was examined in \cite{Neumann2023fast} for the problems under the deterministic settings. 
In our approach, we modify its fitness function, replacing the calculation of deterministic weight with the defined fitness function.
Fundamentally, the \sg shares the same mechanism and mutation operator as the \gs.
However, the \sg introduces a sliding-selection process (see in Algorithm ~\ref{alg:sliding}) to choose some potential individuals from the whole population for mutation.
This sliding-selection defines a window based on the given bound $B$ and the total time $t_{max}$.
The window is sliding according to the current time $t$, with the lower and upper bounds set as $\left\lfloor\frac{tB}{t_{max}}\right\rfloor$ and $\left\lceil\frac{tB}{t_{max}}\right\rceil$, respectively.
As $t$ increases, so does the lower bound of the window.
The algorithm selects an individual for mutation from within the window if there are any with evaluated weight;
if not, it randomly selects from the original population.

\begin{algorithm}[tb]
\raggedright
\caption{GSEMO}
\label{alg:gsemo}
\textbf{Input}: Probability $\alpha$, bound $B$ \\
\textbf{Output}: the best individual $X$
\begin{algorithmic}[1] 
\STATE Set $X = 0^n$;
\STATE $P\gets \{X\}$;
\REPEAT
\STATE Choose $X\in P$ uniformly at random;
\STATE $Y\gets$ flip each bit of $X$ with probability $\frac{1}{n}$;
\IF{$\nexists W \in P : W\succ Y$}
\STATE $P\gets (P\setminus\{Z\in P | Y \succeq Z\})\cup \{Y\}$;
\ENDIF
\UNTIL{stop;}
\end{algorithmic}
\end{algorithm}

\begin{algorithm}[tb]
\raggedright
\caption{SW-GSEMO}
\label{alg:sw_gsemo}
\textbf{Input}: Total time $t_{max}$, probability $\alpha$, bound $B$ \\
\textbf{Output}: the best individual $X$
\begin{algorithmic}[1] 
\STATE Set $X = 0^n$;
\STATE $P\gets \{X\}$;
\STATE $t \gets 0$;
\REPEAT
\STATE $t = t+1$
\STATE Choose $X =$sliding-selection$(P,t,t_{max},\alpha,B)$; \label{line:sw}
\STATE $Y\gets$ flip each bit of $X$ with probability $\frac{1}{n}$;
\IF{$\nexists W \in P : W\succ Y$}
\STATE $P\gets (P\setminus\{Z\in P | Y \succeq Z\})\cup \{Y\}$
\ENDIF
\UNTIL{$t\geq t_{max}$;}
\end{algorithmic}
\end{algorithm}

\begin{algorithm}[tb]
\raggedright
\caption{Sliding-selection}
\label{alg:sliding}
\textbf{Input}: Population P, current iteration $t$, total time $t_{max}$, probability $\alpha$, bound $B$\\
\textbf{Output}: the selected individual $X$
\begin{algorithmic}[1] 
\IF{$t\leq t_{max}$}
\STATE $\hat{c}\gets (t/t_{max})\cdot B$;
\STATE $\widehat{P} = \{X\in P \mid \lfloor\hat{c}\rfloor\leq g_2(X)\leq \lceil\hat{c}\rceil\}$;
\IF{$\widehat{P} = \emptyset$}
\STATE $\widehat{P}\gets P$;
\ENDIF
\ELSE
\STATE $\widehat{P}\gets P$;
\ENDIF
\STATE Choose $X\in \widehat{P}$ uniformly at random;
\STATE \textbf{Return} $X$;

\end{algorithmic}
\end{algorithm}

\section{Fast sampling-based Pareto Optimization}
\label{sec:fspo}
We now introduce our proposed novel algorithm utilizing the sampling-based approach. 
In the sampling-based approach, we independently sample the actual weight of each element from a given uniform distribution $T_{sp}$ times.
This results in a solution $X$ with a set of independent sampling weights, denoted by $Ws$.
Next, we sort the sampling weights in descending order. 
With the $index = \lceil T_{sp}\cdot \alpha \rceil$, the maximal sampling weight $W_{sp}(X)$ that meets the chance constraint under the probability $\alpha$ is obtained, i.e., $W_{sp}(X) = Ws[index]$.
Therefore, we can consider that the solution is feasible if the sampling weight of solution $X$ does not exceed the bound (i.e., $W_{sp}(X)\leq B$); otherwise, the solution is deemed infeasible.
The pseudo-code for this approach is outlined in Algorithm~\ref{alg:sp_eva}.
With the sampling weight $W_{sp}(X)$,
the function to minimize the weight of solution in fitness changes to $g_3(X) = W_{sp}(X)$. 
The new fitness function becomes $g = (g_1(X),g_3(X))$. 
We have $X\succeq Y$ iff $g_1(X)\geq g_1(Y)$ and $g_3(X)\leq g_3(Y)$.


In the novel \asg algorithm, we modify the sliding-selection process in Algorithm~\ref{alg:sw_gsemo} at line~\ref{line:sw} by combining with an adaptive sliding-selection approach. 
This algorithm is a variant of the \sg.
It's important to note that in Algorithm~\ref{alg:sliding}, the window size is fixed at exactly one.
This limitation makes it challenging to capture potential individuals when the weights of the elements are significantly larger than 1.
Consequently, the results yielded by the \sg tend to be similar to those of the \gs when dealing with larger weights. 
To address these challenges, we introduce the adaptive sliding-selection method (described in Algorithm~\ref{alg:ad_sliding}). 
The lower bound of the adaptive window is defined as the same way as the sliding-selection approach,
but a global variable $w_{size}$ is used to control the size of the window.
When no individual is present within the window, $w_{size}$ increases by one. 
However, if more than one individual is in the window and $w_{size}$ is larger than 1, $w_{size}$ decreases by 1; otherwise, $w_{size}$ remains.
Additionally, similar to the classical sliding window approach,
if there are no individuals within the adaptive window, the algorithm selects an individual as a parent from the entire population. 




\begin{algorithm}[tb]
\caption{Sampling weight of solution ($W_{sp}$)}
\label{alg:sp_eva}
\raggedright
\textbf{Input}: Solution $x$, sampling size $T_{sp}$, probability $\alpha$, weight sampling set $\{\hat{w_i}|0\leq i\leq n\}$.\\
\textbf{Output}: the sampling weight
\begin{algorithmic}[1] 
\STATE $t=0$, $Ws\gets \emptyset$, $index = \lceil T_{sp}\cdot \alpha \rceil$;
\WHILE{$t < T_{sp}$}
\STATE Sample $ w_i \sim \hat{w_i}$;
\STATE $w_s = \sum_{i=1}^n w_ix_i$;
\STATE $W_s.\text{add}(w_s)$
\STATE $t = t+1$
\ENDWHILE
\STATE $sort(W_s)$ in descending order; \label{alg_line:sort}
\RETURN $W_s[index]$
\end{algorithmic}
\end{algorithm}

\begin{algorithm}[tb]
\raggedright
\caption{Adaptive sliding-selection}
\label{alg:ad_sliding}
\textbf{Input}: Population P, current iteration $t$, total time $t_{max}$, probability $\alpha$, bound $B$\\
\textbf{Output}: the selected individual $X$
\begin{algorithmic}[1] 
\STATE Global $w_{size}$
\IF{$t\leq t_{max}$}
\STATE $\hat{c}\gets (t/t_{max})\cdot B$;
\STATE $\widehat{P} = \{X\in P \mid \lfloor\hat{c}\rfloor\leq g_3(X)\leq \lfloor\hat{c}\rfloor + w_{size}\}$;
\IF{$\widehat{P} = \emptyset$}
\STATE $\widehat{P}\gets P$;
\STATE $w_{size} = w_{size} + 1$
\ELSIF{$w_{size}> 1$ and $|\widehat{P}| > 1$}
\STATE $w_{size} = w_{size} - 1$
\ENDIF
\ELSE
\STATE $\widehat{P}\gets P$;
\ENDIF
\STATE Choose $X\in \widehat{P}$ uniformly at random;
\STATE \textbf{Return} $X$;

\end{algorithmic}
\end{algorithm}

\section{Experiments}
\label{sec:exp}
In this section, we investigate the performance of \asg and other algorithms that employ the sampling-based evaluation method on the chance-constrained submodular problem across various settings.
We specifically focus on the maximum coverage problem (MCP) based on graphs~\cite{khuller1999budgeted}, which is a kind of classical submodular problem, as part of our experimental work.
Additionally, we conduct a comparison of the results achieved by the \asg against those garnered by the \gs and \sg using the sampling-based evaluation approach. 
Moreover, we examine the performances of \asg across various surrogates to highlight the gap among the different evaluation methods.

\subsection{Experimental Setup}

Within this paper, we examine a chance-constrained version of MCP.
To briefly describe, given an undirected graph $G$ with a set of vertices $V = \{v_1,...,v_n\}$ and  a set of edges $E$,
we denote $N(V')$ the number of vertices of $V'$ and their neighbors in the graph.
The objective of MCP is to find a subset $V'\subseteq V$ so that $N(V')$ is maximized according to the chance constraint with a deterministic bound $B$.
Additionally, a linear function $W: V\to \mathbb{R}^+$ is provided to determine the weight of each vertex.
The MCP under chance constraint can be formulated as 
\begin{equation}
    \argmax_{V'\in V} N(V')~s.t~Pr[W(V')>B]\leq \alpha.
\end{equation}

For our experiments, we utilize three larger sparse graphs: ca-GrQc ($n = 4,158$), ca-HepPh ($n = 11,204$) and ca-Astroph ($n = 17,903$).
We denote $D(v_i)$ the degree of each vertex $v_i$.
In our IID weight setting, each vertex is assigned identical expected weights and dispersion values, set at $a = d = n$.
For the uniform weight setting, we define variable expected weights as $a_i =\frac{(n+D(v_i))^5}{n^4}$ and the identical dispersion as $d = n$.
They differ from previous work where smaller values were used in their experimental setup. 
In this specific setting, the impact of uncertainty closely approximates the expected weight when surrogate evaluation methods are applied.
The algorithms are tested with different bounds $B \in \{\lfloor n^2/2\rfloor, n^2\}$ and probability values $\alpha \in \{0.1,0.001\}$ respectively. 
Different sampling times $T_{sp}$ are considered for $\alpha = 0.1$,
where weights are independently sampled 250, 500, and 1,000 times.
These sampling files are stored to ensure that algorithms can obtain the same feasible solution. 
For $\alpha = 0.001$, we set $T_{sp} = 1000$.
Furthermore, the maximum iterations for the \gs and the total times for the \sg and \asg are set to 1,500,000. 
To statistically compare the performance of the algorithms, 
each one is executed for 30 independent runs, and the minimum, maximum, average values, and standard deviation of the results will be reported.
Additionally, the \asg using surrogates is tested under the same experimental setup.
To verify the differences between the various evaluation methods, we apply the Kruskal-Wallis test with a 95\% confidence level to ensure the statistical validity of the results.

\subsection{The Result on Uniform IID Weights}

\begin{table*}[ht]
\caption{Results for Maximum coverage problem with IID weights using sampling-based evaluation}
\label{table: r_iid}
\resizebox{0.9\textwidth}{4cm}{
\begin{tabular}{@{}lllllllllllllllllll@{}}
\toprule
        &         &          &          & \multicolumn{5}{c}{GESMO}                    & \multicolumn{5}{c}{SW-GESMO}                 & \multicolumn{5}{c}{ASW-GESMO}                          \\ \midrule
Graph   & $B$     & $\alpha$ & $T_{sp}$ & Min   & Max   & Mean      & std     & $|V'|$ & Min   & Max   & Mean      & std     & $|V'|$ & Min   & Max   & {Mean}      & std      & $|V'|$ \\
ca-GrQc    & $n^2/2$ & 0.1      & 250      & 3584  & 3688  & 3644.933  & 22.162  & 787    & 3620  & 3681  & 3647.466  & 14.95   & 789    & 4051  & 4157  & \textbf{4137.733}  & 21.866   & 807    \\
        &         &          & 500      & 3600  & 3671  & 3634.533  & 18.402  & 780    & 3608  & 3680  & 3645.3    & 21.537  & 792    & 3931  & 4157  & \textbf{4126.2}    & 46.314   & 806    \\
        &         &          & 1000     & 3600  & 3685  & 3643.6    & 22.808  & 737    & 3594  & 3687  & 3647.133  & 20.744  & 803    & 3641  & 4155  & \textbf{4109.266}  & 103.252  & 800    \\
        &         & 0.001    & 1000     & 3619  & 3709  & 3656.866  & 23.376  & 836    & 3583  & 3696  & 3646      & 23.613  & 804    & 3903  & 4157  & \textbf{4134.866}  & 46.303   & 810    \\
        & $n^2$   & 0.1      & 250      & 3611  & 3698  & 3650.166  & 22.853  & 794    & 3597  & 3723  & 3648.2    & 26.184  & 791    & 3690  & 4151  & \textbf{4072.366}  & 88.084   & 791    \\
        &         &          & 500      & 3605  & 3684  & 3648.733  & 20.291  & 806    & 3591  & 3685  & 3649.133  & 25.828  & 795    & 3685  & 4145  & \textbf{4040.366}  & 97.905   & 777    \\
        &         &          & 1000     & 3613  & 3695  & 3652.133  & 20.611  & 797    & 3606  & 3678  & 3648.533  & 18.768  & 794    & 3696  & 4142  & \textbf{4047.733}  & 100.77   & 785    \\
        &         & 0.001    & 1000     & 3598  & 3685  & 3649.433  & 23.048  & 806    & 3597  & 3718  & 3653.033  & 25.541  & 802    & 3865  & 4147  & \textbf{4069.3}    & 67.987   & 798    \\ \midrule
ca-HepPh   & $n^2/2$ & 0.1      & 250      & 7907  & 8195  & 8090.5    & 63.93   & 1286   & 7928  & 8216  & 8077.966  & 65.415  & 1026   & 8175  & 10890 & \textbf{10244.8}   & 661.6687 & 1383   \\
        &         &          & 500      & 7943  & 8233  & 8066.433  & 60.02   & 1093   & 7938  & 8230  & 8081.233  & 70.599  & 1072   & 8310  & 11002 & \textbf{10090.766} & 711.786  & 1377   \\
        &         &          & 1000     & 7941  & 8178  & 8055.1    & 54.066  & 1148   & 7936  & 8222  & 8077.1    & 58.3    & 1065   & 8695  & 11043 & \textbf{10152.266} & 518.776  & 1381   \\
        &         & 0.001    & 1000     & 7982  & 8179  & 8083.933  & 54.247  & 1056   & 7943  & 8179  & 8054.866  & 45.394  & 1033   & 8603  & 10966 & \textbf{10184.9}   & 666.218  & 1296   \\
        & $n^2$   & 0.1      & 250      & 7902  & 8142  & 8052.766  & 67.183  & 1047   & 7951  & 8205  & 8071.433  & 64.906  & 1073   & 8122  & 10737 & \textbf{9707.033}  & 684.183  & 1262   \\
        &         &          & 500      & 8003  & 8134  & 8080.8    & 35.424  & 1050   & 7969  & 8133  & 8050.8    & 45.735  & 1010   & 8239  & 10595 & \textbf{9667.1}    & 633.359  & 1268   \\
        &         &          & 1000     & 7908  & 8244  & 8066.866  & 77.486  & 1032   & 7952  & 8180  & 8067.033  & 54.643  & 1025   & 8129  & 10560 & \textbf{9733.1}    & 566.416  & 1287   \\
        &         & 0.001    & 1000     & 7931  & 8195  & 8080.333  & 59.6    & 1057   & 7911  & 8206  & 8068.033  & 73.88   & 1070   & 8142  & 10593 & \textbf{9614.966}  & 741.976  & 1227   \\ \midrule
ca-AstroPh & $n^2/2$ & 0.1      & 250      & 12498 & 12842 & 12651.033 & 72.806  & 1056   & 12400 & 12699 & 12586.2   & 77.693  & 1027   & 14032 & 17261 & \textbf{16061.6}   & 942.73   & 1550   \\
        &         &          & 500      & 12349 & 12850 & 12617.766 & 88.295  & 1040   & 12458 & 12780 & 12636.4   & 80.161  & 1047   & 13029 & 17063 & \textbf{15640.933} & 1094.443 & 1527   \\
        &         &          & 1000     & 12385 & 12797 & 12634.166 & 102.409 & 1041   & 12502 & 12823 & 12647.766 & 100.1   & 1053   & 12789 & 17440 & \textbf{15873.633} & 1091.795 & 1542   \\
        &         & 0.001    & 1000     & 12399 & 12786 & 12627.633 & 93.628  & 1084   & 12487 & 12892 & 12656.133 & 81.645  & 1090   & 13199 & 17288 & \textbf{15849.866} & 957.862  & 1545   \\
        & $n^2$   & 0.1      & 250      & 12486 & 12903 & 12637.333 & 87.959  & 1104   & 12480 & 12844 & 12657.266 & 80.585  & 1084   & 12744 & 16524 & \textbf{14889.333} & 1026.238 & 1520   \\
        &         &          & 500      & 12482 & 12781 & 12653.966 & 81.374  & 1081   & 12395 & 12870 & 12628.133 & 107.551 & 1061   & 12645 & 16242 & \textbf{14773.66}  & 1063.271 & 1499   \\
        &         &          & 1000     & 12463 & 12812 & 12624.466 & 86.244  & 1068   & 12522 & 12814 & 12670.933 & 79.292  & 1052   & 12505 & 16241 & \textbf{14854.633} & 947.947  & 1507   \\
        &         & 0.001    & 1000     & 12378 & 12771 & 12625.333 & 94.793  & 1059   & 12682 & 16435 & 12650.733 & 93.281  & 1070   & 12457 & 17872 & \textbf{15214.1}   & 900.65   & 1513   \\ \bottomrule
\end{tabular}
}
\end{table*}

\begin{figure*}
    \centering
    \begin{subfigure}[b]{0.49\textwidth}
        \centering
        \includegraphics[scale=0.2]{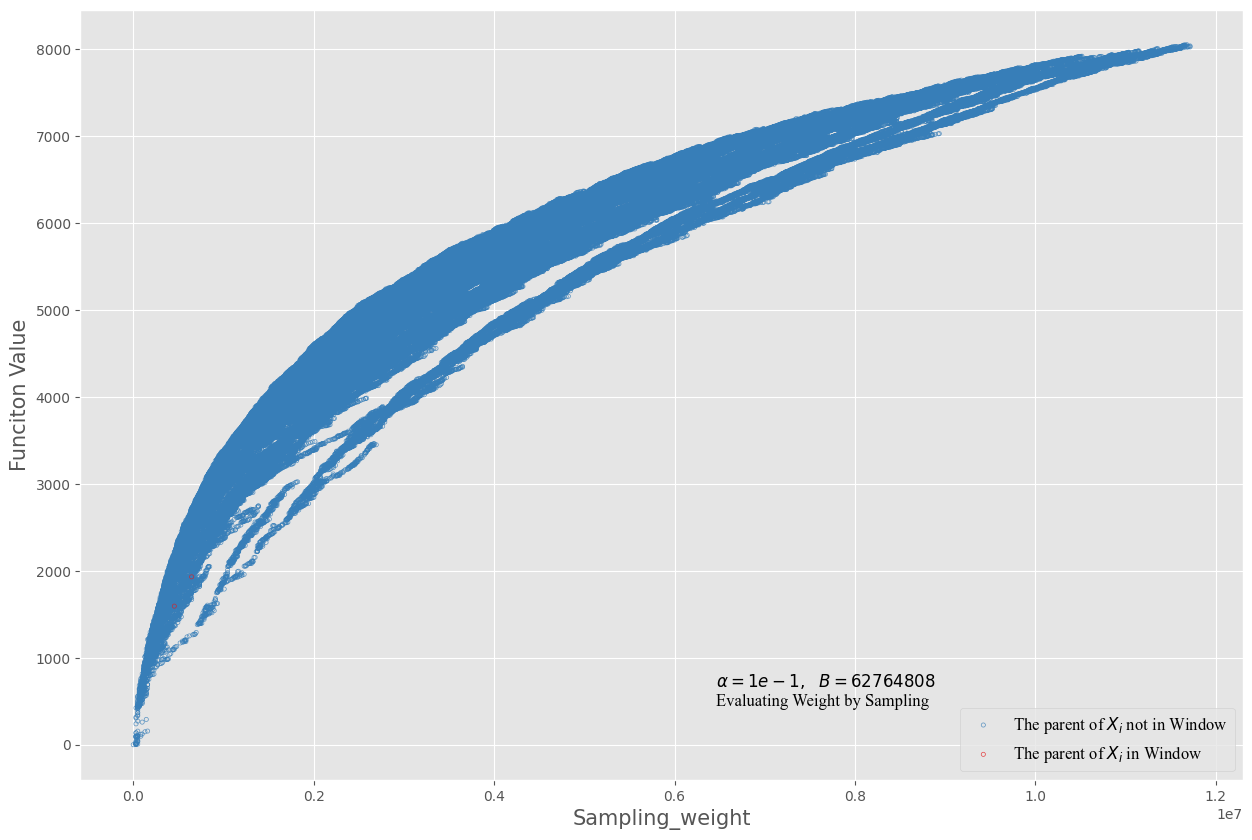}
        \caption{\sg }
        \label{fig:sw_iid}
    \end{subfigure}
    \hfill
    \begin{subfigure}[b]{0.49\textwidth}
        \centering
        \includegraphics[scale=0.2]{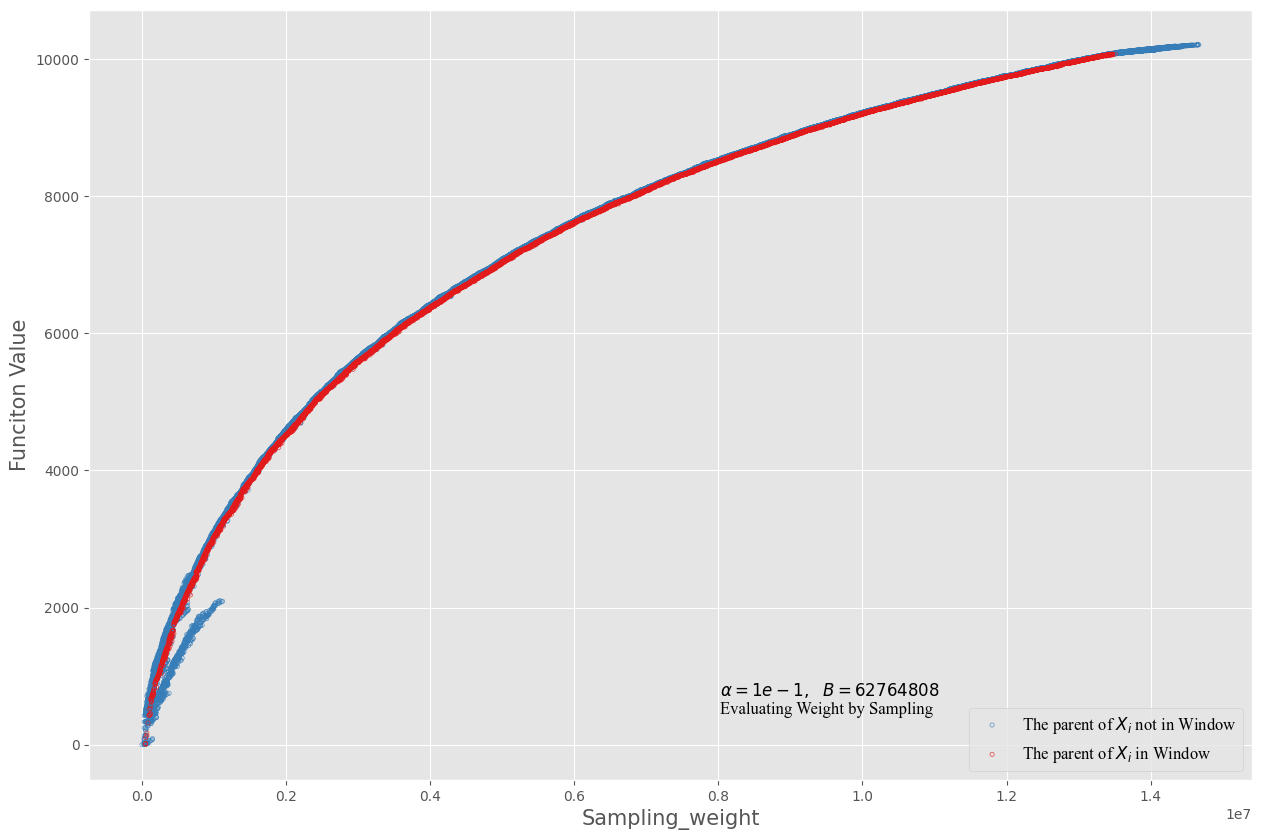}
        \caption{\asg}
        \label{fig:asw_iid}
    \end{subfigure}
    \label{fig:enter-label}
    \caption{Optimization process for ca-HepPh with IID weights using sampling-based evaluation}
\end{figure*}

Here, we focus on the performance of various algorithms on the MCP with IID weights. 
We present the results obtained by the algorithms in Table~\ref{table: r_iid} and compare the performance across different instances. 
Further, we visualize the optimization processing of the \asg to demonstrate its advantage contract to the \sg. 
Additionally, we provide a visualization of the optimization process of the \asg to highlight its advantages over the \sg.
Additionally, we present results from the \asg using the surrogate evaluation method and identify the differences between the two evaluation approaches.

\subsubsection{Results comparison and Visualization of ASW-GSEMO}

\begin{figure*}[ht]
    \centering
    \begin{subfigure}[b]{0.49\textwidth}
        \centering
        \includegraphics[scale=0.2]{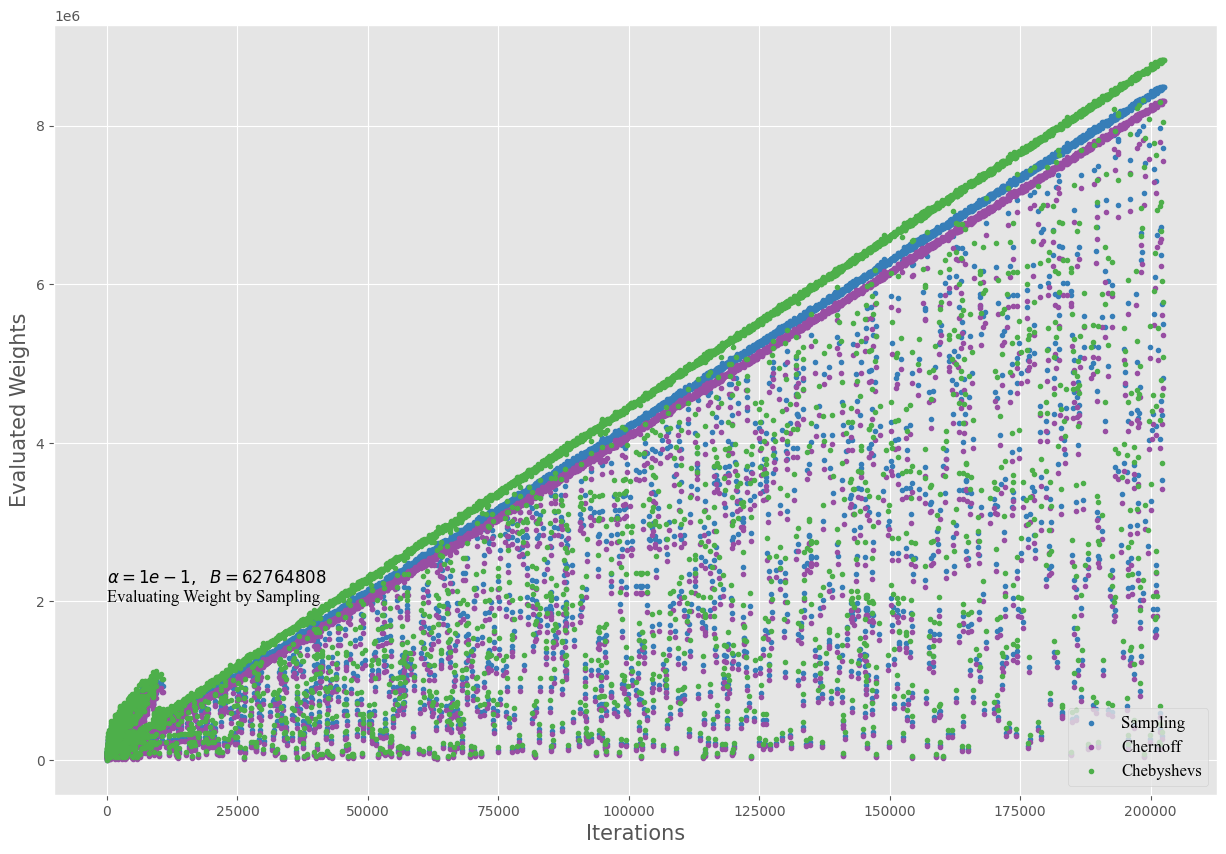}
        \caption{$\alpha = 1e-1$}
        \label{fig:sp_vs_sg_l_iid}
    \end{subfigure}
    \hfill
    \begin{subfigure}[b]{0.49\textwidth}
        \centering
        \includegraphics[scale=0.2]{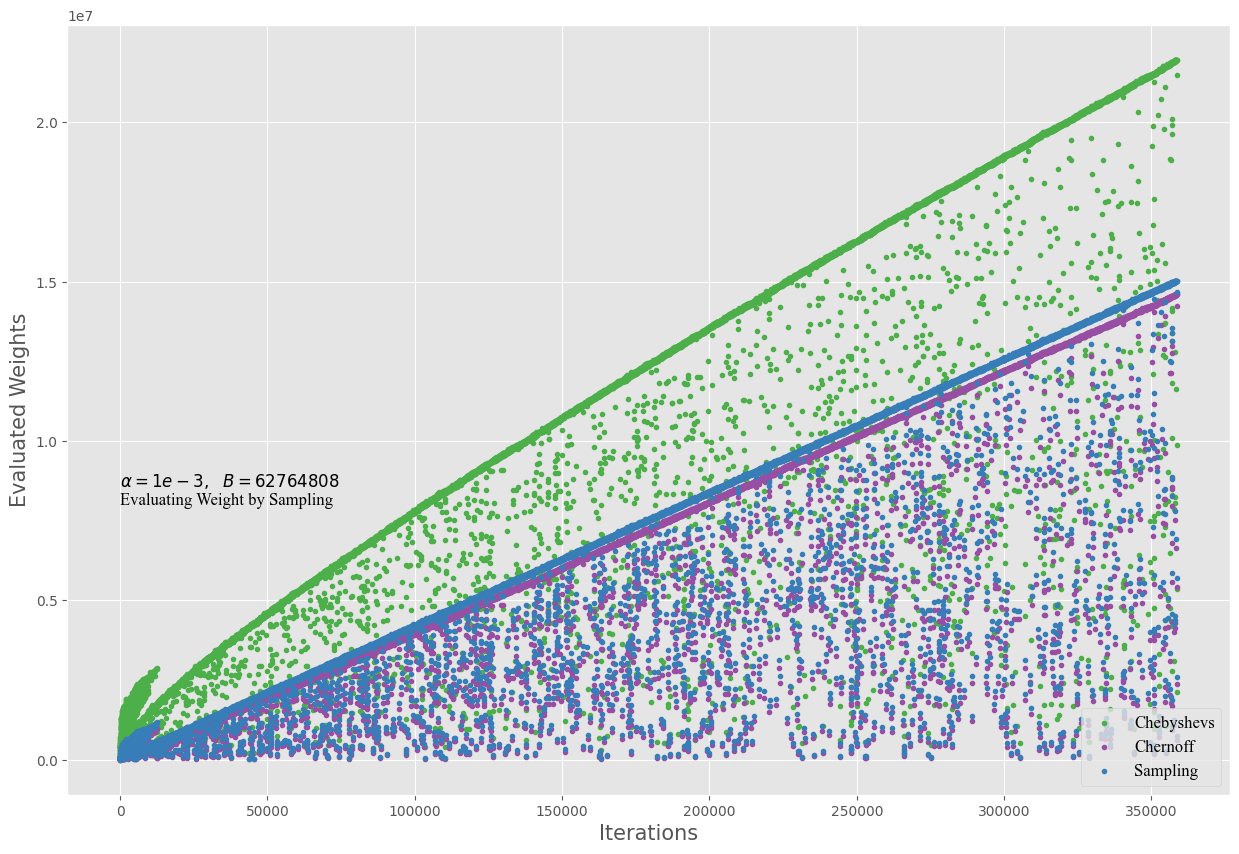}
        \caption{$\alpha = 1e-3$}
        \label{fig:sp_vs_sg_s_iid}
    \end{subfigure}
    \caption{The weights obtained by \asg using different approaches in ca-HepPh with IID weights}
\end{figure*}

The results in Table~\ref{table: r_iid} illustrate that the \asg generally outperforms the \gs and \sg.
However, it is noteworthy that the standard deviation of the \asg is higher than the others across all instances. 
This is because the size adaptive window in different runs is not stable. 
The results also reveal that the performance of the \sg is almost on par with the \gs.
Furthermore, the table indicates that the average number of elements $|V'|$ in the best solution is larger when the \asg is applied,
suggesting its effectiveness in finding more comprehensive solutions.

We also note that the sampling size $T_{sp}$ does not significantly impact the results of the algorithms,
and there is minimal difference in the outcomes for different values of $\alpha$.
This observation suggests that the influence of uncertainty is reduced when the dispersion is as large as the expected weight.
However, in the case of the \asg, the results and the value of $|V'|$ obtained with the larger bound are slightly lower than those from the smaller bound,
indicating that the \asg may not fully converge within 1.5 million iterations.

Furthermore, Table~\ref{table:pop_iid} displays the average population sizes obtained by the algorithms in the graph ca-Astroph.
It reveals that the population size of the \asg is significantly larger than that of the other algorithms when the graph becomes larger, 
and the population size from the \sg is very similar to that of the \gs. 
This finding supports the notion that the performance of the \asg is better than others and the \sg struggles to achieve better results than the \gs when the expected weight is larger.

\begin{table}[ht]
\caption{Average number of trade-off solutions obtained by different algorithms in ca-AstroPh with IID weights}
\label{table:pop_iid}
\resizebox{0.45\textwidth}{1.5cm}{
\begin{tabular}{@{}lcccccc@{}}
\toprule
Graph   & $B$     & $\alpha$ & $T_{sp}$ & GESMO & SW-GESMO & ASW-GESMO \\ \midrule
ca-AstroPh & $n^2/2$ & 0.1      & 250      & 1819  & 1826     & 3334      \\
        &         &          & 500      & 1822  & 1860     & 3402      \\
        &         &          & 1000     & 1838  & 1823     & 3373      \\
        &         & 0.001    & 1000     & 1831  & 1835     & 3294      \\
        & $n^2$   & 0.1      & 250      & 1845  & 1822     & 2971      \\
        &         &          & 500      & 1821  & 1844     & 2865      \\
        &         &          & 1000     & 1836  & 1832     & 2902      \\
        &         & 0.001    & 1000     & 1840  & 1847     & 2917      \\ \bottomrule
\end{tabular}
}
\end{table}

\begin{table}[ht]
\caption{Results for MCP with IID weights using surrogates}
\label{table:sg_iid}
\resizebox{0.47\textwidth}{2.2cm}{
\begin{tabular}{@{}lllllllll@{}}
\toprule
        &         &          & \multicolumn{3}{l}{\asg with $W_{cheb}$} & \multicolumn{3}{l}{\asg with $W_{chen}$} \\ \midrule
Graph   & $B$     & $\alpha$ & mean               & std              & p-value         & mean                & std             & p-value         \\
ca-GrQc    & $n^2/2$ & 0.1      & 4137.9             & 19.871           & 0.314           & 4138.8              & 16.383          & 0.51            \\
        &         & 0.001    & 4137.933           & 17.914           & 0.332           & 4138.3              & 23.629          & 0.505           \\
        & $n^2$   & 0.1      & 4122.433           & 14.247           & 0.0002          & 4122.133            & 17.753          & 0.0002          \\
        &         & 0.001    & 4130.366           & 19.062           & 4.191           & 4118.3              & 23.179          & 0.004           \\ \midrule
ca-HepPh   & $n^2/2$ & 0.1      & 10844.133          & 205.511          & 0               & 10851.4333          & 226.556         & 0               \\
        &         & 0.001    & 10926.833          & 122.11           & 0               & 10839.6             & 189.793         & 0               \\
        & $n^2$   & 0.1      & 10547.133          & 242.735          & 0               & 10364.033           & 406.008         & 0               \\
        &         & 0.001    & 10715.266          & 253.327          & 0               & 10551.5             & 305.352         & 0               \\ \midrule
ca-AstroPh & $n^2/2$ & 0.1      & 17058.3            & 324.184          & 0               & 16759.1             & 554.048         & 0               \\
        &         & 0.001    & 17183.4            & 180.035          & 0               & 17061.1             & 412.126         & 0               \\
        & $n^2$   & 0.1      & 16146.333          & 622.756          & 0               & 15978.4             & 715.088         & 0               \\
        &         & 0.001    & 16594.866          & 322.304          & 0               & 16074               & 623.282         & 0               \\ \bottomrule
\end{tabular}
}
\end{table}

\begin{table}[ht]
\caption{ Final Population size and average of $|V'|$  for MCP with IID weights using Surrogates}
\label{table:pav_sg_iid}
\resizebox{0.47\textwidth}{2.2cm}{
\begin{tabular}{@{}lllcccc@{}}
\toprule
        &         &          & \multicolumn{2}{c}{\asg with $W_{cheb}$} & \multicolumn{2}{c}{\asg with $W_{chen}$} \\ \midrule
Graph   & $B$     & $\alpha$ & pop\_size                    & $|V'|$                   & pop\_size                    & $|V'|$                   \\
ca-GrQc    & $n^2/2$ & 0.1      & 796                          & 801                      & 791                          & 799                      \\
        &         & 0.001    & 792                          & 797                      & 796                          & 795                      \\
        & $n^2$   & 0.1      & 783                          & 793                      & 793                          & 801                      \\
        &         & 0.001    & 786                          & 808                      & 793                          & 804                      \\ \midrule
ca-HepPh   & $n^2/2$ & 0.1      & 1555                         & 1517                     & 1522                         & 1536                     \\
        &         & 0.001    & 1532                         & 1559                     & 1566                         & 1570                     \\
        & $n^2$   & 0.1      & 1509                         & 1563                     & 1480                         & 1593                     \\
        &         & 0.001    & 1422                         & 1496                     & 1404                         & 1478                     \\ \midrule
ca-AstroPh & $n^2/2$ & 0.1      & 1650                         & 1756                     & 1587                         & 1748                     \\
        &         & 0.001    & 1796                         & 1836                     & 1740                         & 1749                     \\
        & $n^2$   & 0.1      & 1789                         & 1755                     & 1610                         & 1708                     \\
        &         & 0.001    & 1680                         & 1780                     & 1754                         & 1737                     \\ \bottomrule
\end{tabular}
}
\end{table}

\begin{table*}[ht]
\caption{Results for the maximum coverage problem with uniform weights using sampling-based evaluation
}
\label{table: r_uw}
\resizebox{0.9\textwidth}{4cm}{
\begin{tabular}{@{}lllllllllllllllllll@{}}
\toprule
        &         &          &          & \multicolumn{5}{c}{GESMO}                  & \multicolumn{5}{c}{SW-GESMO}              & \multicolumn{5}{c}{ASW-GESMO}                        \\ \midrule
Graph   & $B$     & $\alpha$ & $T_{sp}$ & Min   & Max   & Mean      & std     & |V'| & Min   & Max   & Mean      & std    & |V'| & Min   & Max   & Mean               & std      & |V'| \\
ca-GrQc    & $n^2/2$ & 0.1      & 250      & 3595  & 3685  & 3633.766  & 24.418  & 828  & 3570  & 3694  & 3638.366  & 27.451 & 823  & 3996  & 4158  & \textbf{4131.2}    & 38.78    & 810  \\
        &         &          & 500      & 3568  & 3688  & 3636.4    & 25.197  & 835  & 3607  & 3677  & 3643.166  & 21.069 & 797  & 3813  & 4156  & \textbf{4121.8}    & 65.746   & 819  \\
        &         &          & 1000     & 3594  & 3668  & 3628.7    & 19.645  & 818  & 3552  & 3684  & 3622.233  & 25.422 & 790  & 3907  & 4155  & \textbf{4110.766}  & 72.816   & 819  \\
        &         & 0.001    & 1000     & 3599  & 3714  & 3635.366  & 25.293  & 809  & 3593  & 3687  & 3634.666  & 23.328 & 808  & 3880  & 4157  & \textbf{4133.366}  & 49.421   & 817  \\
        & $n^2$   & 0.1      & 250      & 3588  & 3685  & 3638.666  & 24.57   & 811  & 3605  & 3662  & 3634      & 16.923 & 808  & 3843  & 4150  & \textbf{4072.466}  & 59.972   & 805  \\
        &         &          & 500      & 3582  & 3706  & 3640.2    & 25.511  & 814  & 3595  & 3668  & 3632.466  & 21.3   & 794  & 3793  & 4141  & \textbf{4064.533}  & 73.632   & 806  \\
        &         &          & 1000     & 3596  & 3676  & 3633.333  & 19.082  & 822  & 3596  & 3656  & 3628.066  & 18.467 & 790  & 3813  & 4147  & \textbf{4053.533}  & 94.099   & 813  \\
        &         & 0.001    & 1000     & 3585  & 3677  & 3634.3    & 24.822  & 817  & 3582  & 3686  & 3629.466  & 22.192 & 789  & 3854  & 4145  & \textbf{4071.2}    & 66.501   & 815  \\ \midrule
ca-HepPh   & $n^2/2$ & 0.1      & 250      & 7929  & 8154  & 8022.7    & 61.173  & 983  & 7957  & 8156  & 8053.933  & 58.424 & 986  & 9235  & 10978 & \textbf{10441.6}   & 412.405  & 1382 \\
        &         &          & 500      & 7940  & 8116  & 8028.6    & 50.631  & 995  & 7921  & 8127  & 8012.066  & 52.695 & 991  & 8173  & 11004 & \textbf{10065}     & 672.44   & 1336 \\
        &         &          & 1000     & 7843  & 8210  & 8031.266  & 69.873  & 997  & 7869  & 8071  & 8000.333  & 55.976 & 995  & 8079  & 11038 & \textbf{10206.266} & 724.934  & 1383 \\
        &         & 0.001    & 1000     & 7871  & 8103  & 8015.333  & 56.041  & 1002 & 7861  & 8155  & 8028.333  & 75.37  & 1004 & 8554  & 10967 & \textbf{10398.4}   & 505.33   & 1354 \\
        & $n^2$   & 0.1      & 250      & 7894  & 8158  & 8021      & 59.575  & 998  & 7936  & 8111  & 8009.066  & 45.386 & 993  & 8297  & 10572 & \textbf{9596.766}  & 688.814  & 1390 \\
        &         &          & 500      & 7910  & 8129  & 8024.7    & 52.235  & 997  & 7945  & 8146  & 8050.433  & 46.745 & 999  & 8174  & 10539 & \textbf{9604.633}  & 712.598  & 1432 \\
        &         &          & 1000     & 7957  & 8152  & 8052.633  & 57.413  & 986  & 7927  & 8180  & 8029.3    & 58.182 & 994  & 8118  & 10527 & \textbf{9636.133}  & 682.395  & 1381 \\
        &         & 0.001    & 1000     & 7890  & 8197  & 8031.8    & 66.824  & 996  & 7927  & 8180  & 8029.3    & 58.182 & 994  & 8224  & 10594 & \textbf{9850}      & 709.575  & 1358 \\ \midrule
ca-AstroPh & $n^2/2$ & 0.1      & 250      & 12428 & 12763 & 12580.533 & 86.087  & 1043 & 12506 & 12728 & 12605.4   & 58.133 & 1038 & 12856 & 17055 & \textbf{16003.333} & 970.202  & 1652 \\
        &         &          & 500      & 12446 & 12785 & 12593.433 & 76.587  & 1052 & 12448 & 12823 & 12579.8   & 97.399 & 1072 & 12830 & 17239 & \textbf{15898.2}   & 1181.282 & 1672 \\
        &         &          & 1000     & 12362 & 12750 & 12541.833 & 85.384  & 1041 & 12373 & 12725 & 12560.566 & 88.95  & 1044 & 12890 & 17262 & \textbf{15334.866} & 1356.371 & 1510 \\
        &         & 0.001    & 1000     & 12378 & 12758 & 12584.733 & 93.7    & 1049 & 12360 & 12732 & 12570.066 & 96.854 & 1059 & 12522 & 17359 & \textbf{15548.466} & 1262.312 & 1536 \\
        & $n^2$   & 0.1      & 250      & 12412 & 12763 & 12588.366 & 80.192  & 1041 & 12479 & 12828 & 12633.033 & 79.822 & 1048 & 12858 & 16796 & \textbf{15408.966} & 875.922  & 1656 \\
        &         &          & 500      & 12433 & 12725 & 12574.733 & 86.624  & 1040 & 12455 & 12838 & 12599.533 & 94.084 & 1030 & 12744 & 16484 & \textbf{14670.4}   & 1025.452 & 1420 \\
        &         &          & 1000     & 12373 & 12922 & 12599.466 & 101.935 & 1051 & 12365 & 12741 & 12562.566 & 93.338 & 1042 & 12653 & 16377 & \textbf{14897.533} & 1045.866 & 1450 \\
        &         & 0.001    & 1000     & 12446 & 12763 & 12565.233 & 81.569  & 1040 & 12423 & 12838 & 12601.166 & 98.533 & 1053 & 12593 & 16572 & \textbf{15047.433} & 1109.39  & 1506 \\ \bottomrule
\end{tabular}
}
\end{table*}


Figures \ref{fig:sw_iid} and \ref{fig:asw_iid} illustrate the optimization processes of the \sg and \asg, respectively, in the graph ca-HepPh. 
They present the relationship between the sampling-based weight and the function value of the solution selected into the population. 
The solutions are labeled to show whether their parents are within the corresponding windows of the algorithms. 
It is evident that the \sg has difficulty capturing individuals within the sliding window,
resulting in many search points positioned far from the Pareto front. 
Conversely, after the \asg adjusts the window length at the start, it captures more individuals in the enhanced window. 
Consequently, most of the solutions generated by the \asg are closer to the Pareto front, 
which provides evidence of why the performance of the \asg is better.




\subsubsection{Sampling VS. Surrogate}

The results obtained from the \asg algorithm using surrogate functions are presented in Table~\ref{table:sg_iid}.
This table also includes statistical p-values comparing the sampling-based approach with two different surrogate approaches respectively. 
For the ca-Grac graph, which has a smaller number of nodes and a lower bound, the p-values are less than 0.05. 
This suggests that the performance of the \asg with the sampling method is comparable to that with the surrogates. 
However, for larger graphs and bounds, the performance of the \asg with the sampling method is not as good as to that with the surrogates.
These findings indicate that while the algorithm with the sampling-based evaluation is as effective as the surrogate methods in instances with smaller expected weights and tighter bounds,
it slightly does not perform as efficiently compared to the surrogates in cases with larger bounds and graphs, especially with fewer iterations.
Table~\ref{table:pav_sg_iid} in the appendix displays the average population size and the average number of elements in the final best solution across different instances. 
According to Tables \ref{table: r_iid} and \ref{table:pav_sg_iid}, the \asg with the sampling-based approach tends to yield more individuals but fewer elements in the final best solution.

The surrogate weights of solutions in the population obtained from the sampling-based method in each iteration are calculated, 
and a portion of these results are visualized in Figures~\ref{fig:sp_vs_sg_l_iid} and \ref{fig:sp_vs_sg_s_iid}.
These figures reveal that while the weights evaluated by different methods are similar when the probability is larger, 
there is a significant discrepancy between the weights evaluated by the surrogate based on one-sided Chebyshev's inequality and other approaches when the probability is smaller.
Additionally, they indicate that the evaluated weights of the solutions do not reach the bound, suggesting incomplete convergence.
The trend of increasing weights implies that the weights evaluated by the surrogate based on one-sided Chebyshev's inequality would likely reach the bound first upon full convergence.


\subsection{The Result on Uniform Weights with Same Dispersion}

\begin{figure*}
    \centering
    \begin{subfigure}[b]{0.49\textwidth}
        \centering
        \includegraphics[scale=0.25]{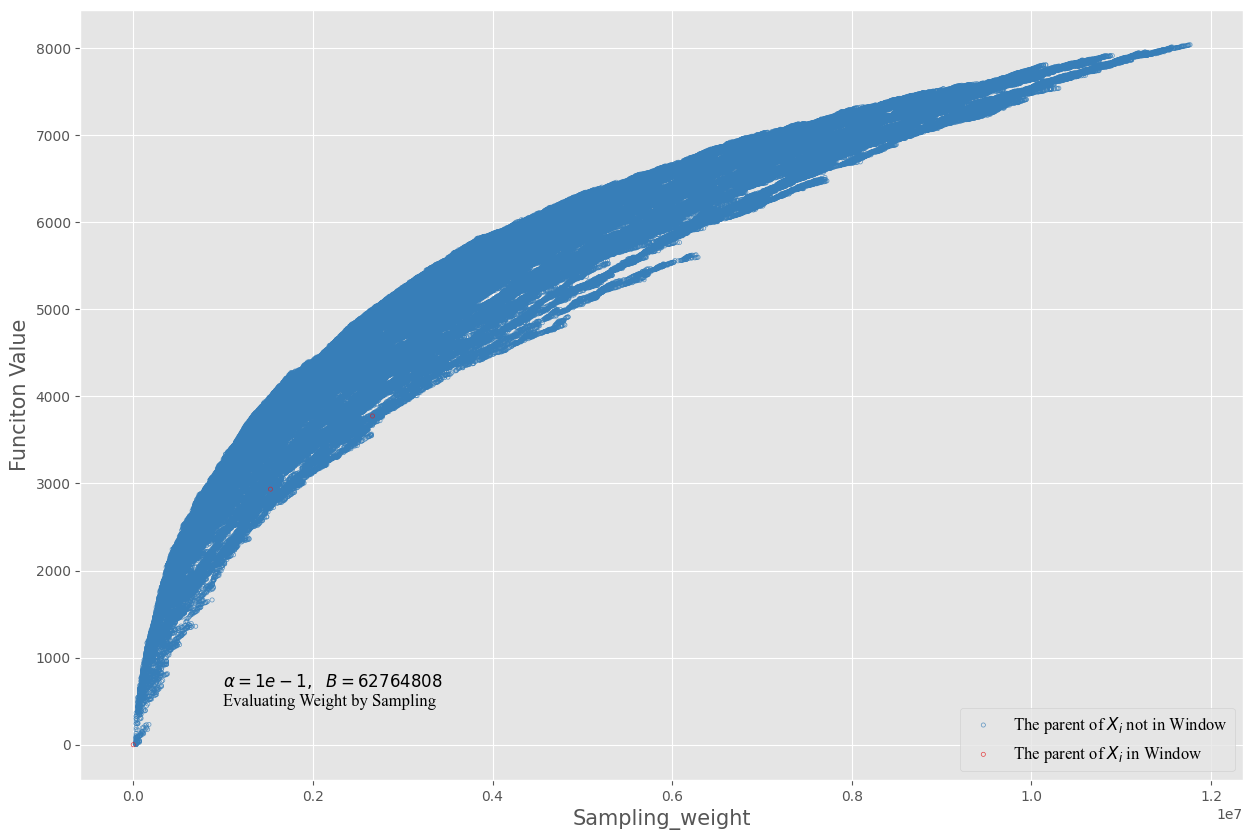}
        \caption{\sg }
        \label{fig:sw_uw}
    \end{subfigure}
    \hfill
    \begin{subfigure}[b]{0.49\textwidth}
        \centering
        \includegraphics[scale=0.25]{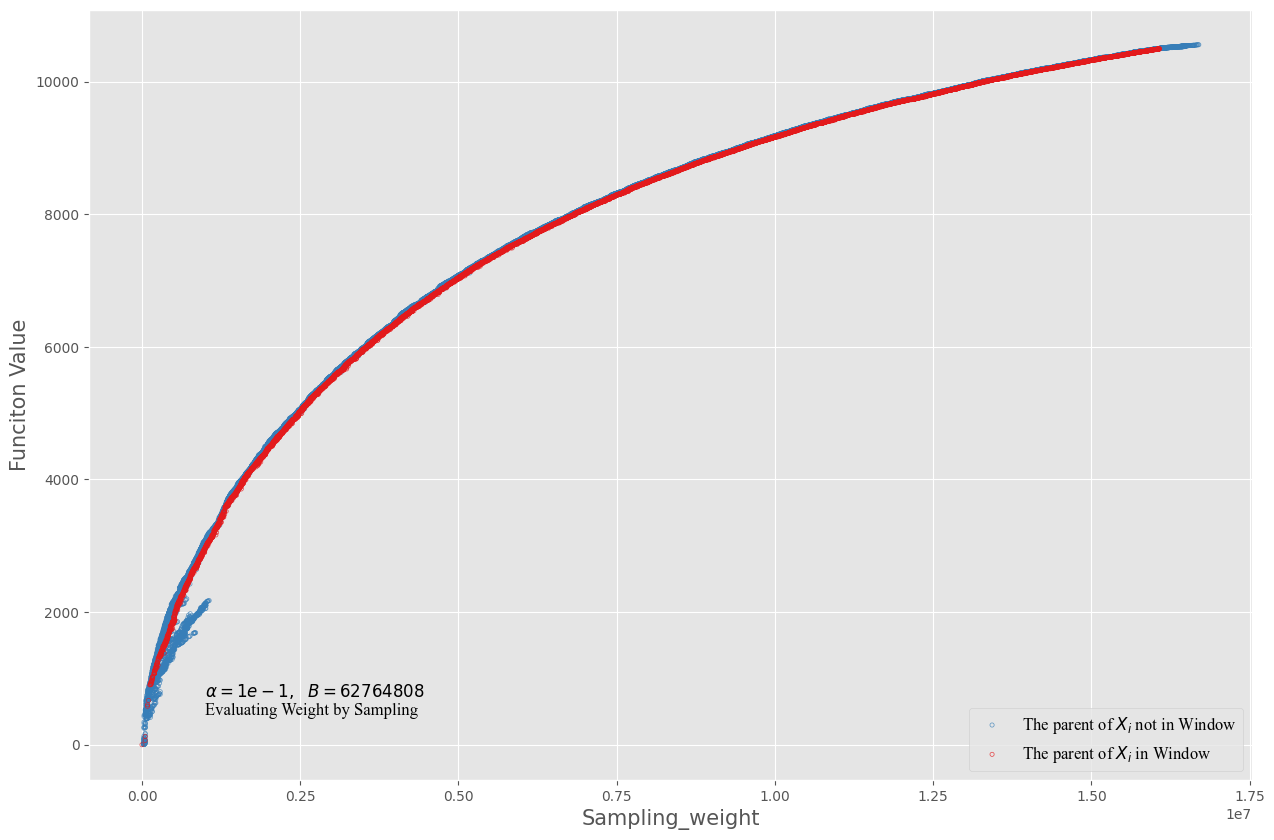}
        \caption{\asg}
        \label{fig:asw_uw}
    \end{subfigure}
    \caption{Optimization process for ca-HepPh with uniform weights using sampling-based evaluation
    }
\end{figure*}

Here, we examine the performance of the \asg on MCP under uniform weights with the same dispersion, comparing it against to other fast Pareto optimization algorithms.
We also visualize the process of the \asg to demonstrate the functioning of its window mechanism.
Additionally, we investigate the differences between various methods of evaluating chance constraints.

\subsubsection{Results comparison and Visualization of ASW-GSWMO}

Table~\ref{table: r_uw} presents the results of different algorithms, including the average number of elements in the best solution.
These findings reflect the same outcomes observed when the algorithm was tested on a problem featuring IID weights.
In this setting, the \asg selects a greater number of elements for the solution, beating the performance of other algorithms. 
This superior performance is noted despite a higher standard deviation, which results from the unstable size of the adaptive window.
The results also show that the sampling size $T_{sp}$ does not significantly affect the final outcomes for both algorithms,
and the disparity between results for different values of $\alpha$ is small.

Figures~\ref{fig:sw_uw} and \ref{fig:asw_uw} are visualized to represent the correlation between sampling-based weights and the corresponding values of the function. 
Analysis of Figures~\ref{fig:sw_uw} and \ref{fig:asw_uw} reveals that the \sg faces challenges in incorporating solutions into the window during the optimization process. 
In contrast, the \asg performs effectively, successfully capturing a larger number of individuals proximate to the Pareto front.
However, it is also observed that the \asg struggles to include certain individuals when the weights reach a certain level in later iterations.
This challenge originates from the initial weights of the solutions not meeting the lower bound of the window, 
complicating the process for the escalating weights to align with the window as the iterations progress.
With additional iterations, these challenges can be mitigated and potentially resolved.

\subsubsection{Sampling VS. Surrogate}

\begin{figure*}[ht]
    \centering
    \begin{subfigure}[b]{0.49\textwidth}
        \centering
        \includegraphics[scale=0.25]{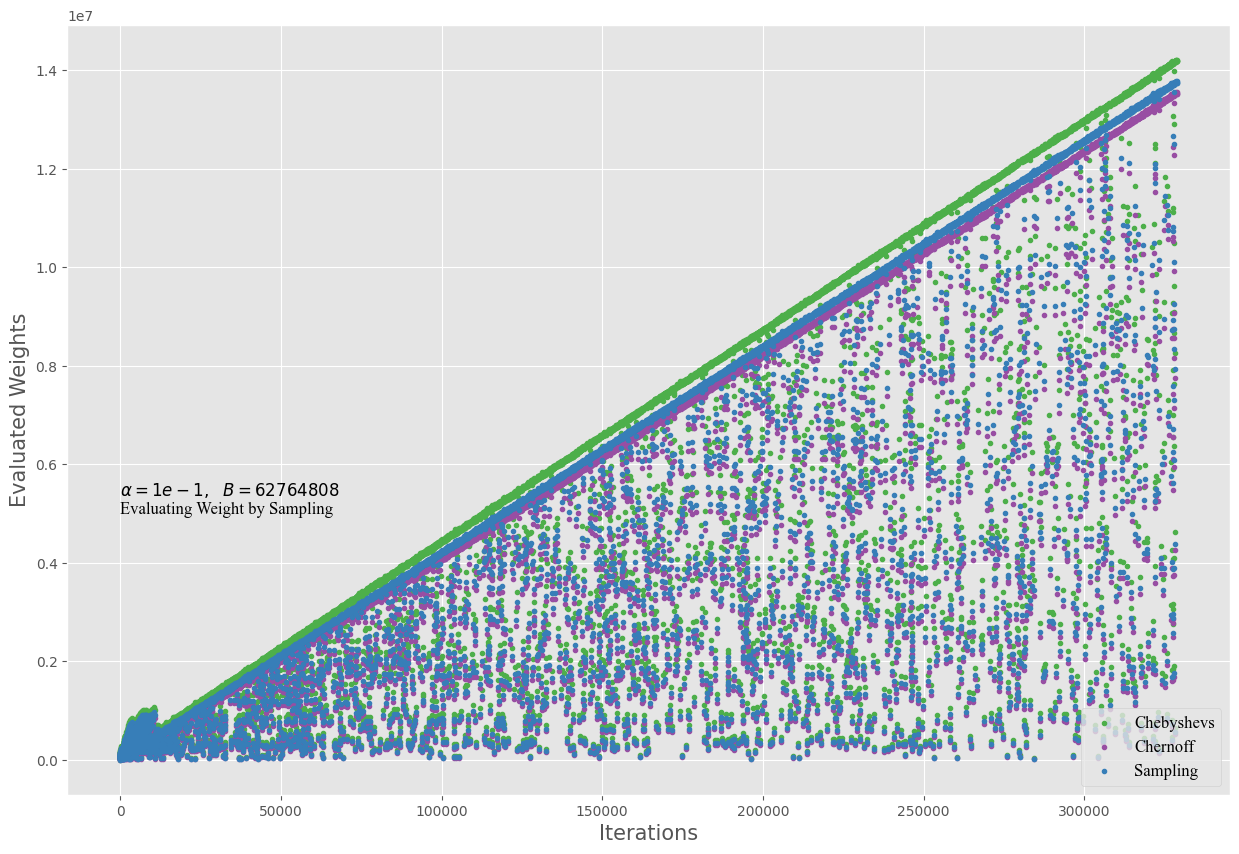}
        \caption{$\alpha = 1e-1$}
        \label{fig:sp_vs_sg_l_uw}
    \end{subfigure}
    \hfill
    \begin{subfigure}[b]{0.49\textwidth}
        \centering
        \includegraphics[scale=0.25]{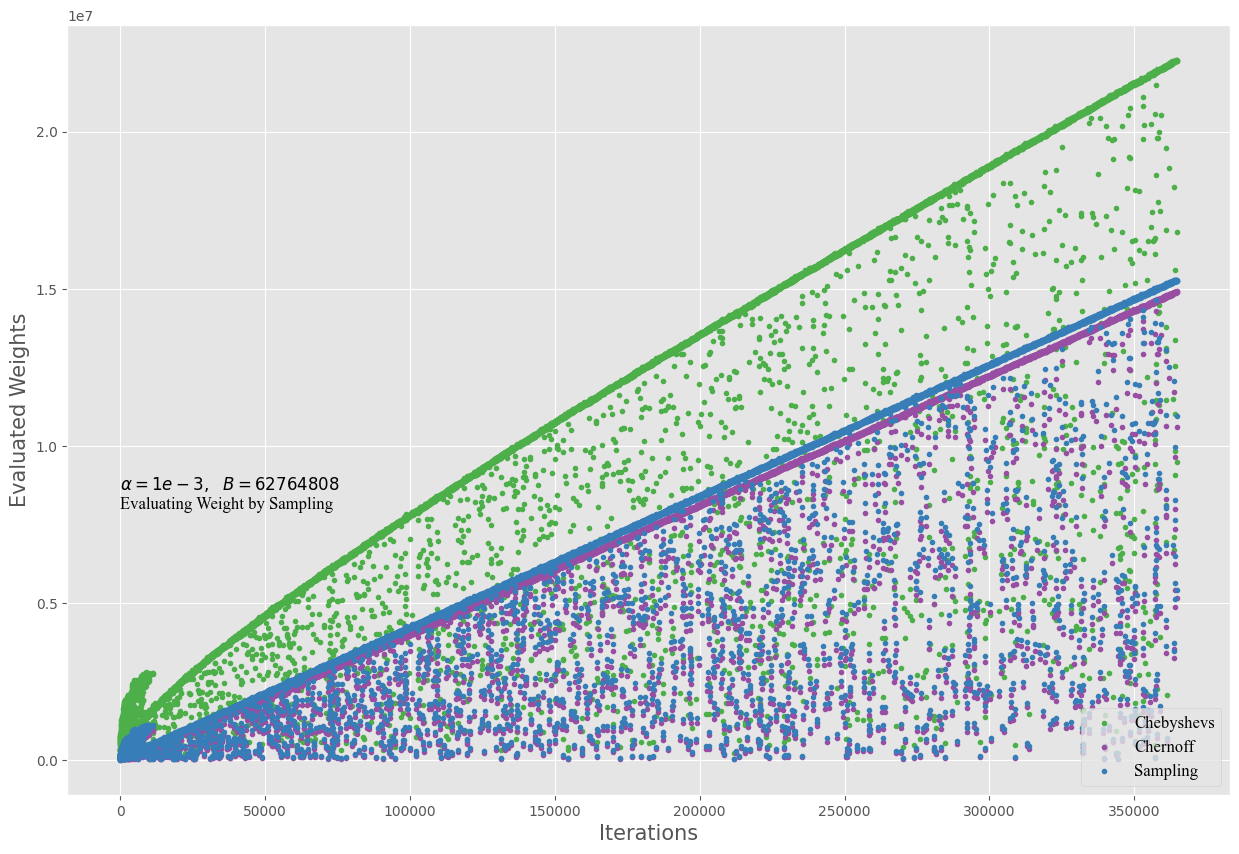}
        \caption{$\alpha = 1e-3$}
        \label{fig:sp_vs_sg_s_uw}
    \end{subfigure}
    \caption{The weights obtained by \asg using different evaluation approaches in ca-HepPh with uniform weights}
\end{figure*}

Table~\ref{table:sg_uw} presents the outcomes and the average number of elements in the best solutions produced by the \asg.
In addition, it includes calculated p-values comparing the results obtained from surrogate and sampling-based methods. 
These results indicate that the performance of the algorithm using the sampling-based approach is as same as that of the surrogate method, 
as evidenced by the fact that the p-values for most cases are below 0.05. 
Furthermore, it is observed that the \asg employing the surrogate method yields solutions with fewer elements compared to those generated using the sampling-based approach.
The influence of variations in $\alpha$ on performance is also minimal, with the results from different surrogate models showing little difference.

\begin{table}[ht]
\caption{Results for MCP with uniform weights using surrogates
}
\label{table:sg_uw}
\resizebox{0.47\textwidth}{2cm}{
\begin{tabular}{@{}lllllllllll@{}}
\toprule
        &         &          & \multicolumn{4}{c}{\asg with $W_{cheb}$} & \multicolumn{4}{c}{\asg with $W_{chen}$} \\ \midrule
Graph   & $B$     & $\alpha$ & mean          & std          & p-value     & $|V'|$     & mean          & std          & p-value     & $|V'|$     \\
ca-GrQc    & $n^2/2$ & 0.1      & 4118.8        & 105.604      & 0.08        & 832        & 4119.7        & 75.798       & 0.06        & 823        \\
        &         & 0.001    & 4135.933      & 62.66        & 0.01        & 828        & 4128.533      & 63.671       & 0.47        & 829        \\
        & $n^2$   & 0.1      & 4023.1        & 113.816      & 0.49        & 810        & 3975.466      & 143.403      & 0.04        & 812        \\
        &         & 0.001    & 4011.333      & 126.258      & 0.29        & 809        & 4010.4        & 101.057      & 0.3         & 812        \\ \midrule
ca-HepPh   & $n^2/2$ & 0.1      & 9863.866      & 659.327      & 0.012       & 1226       & 10068.233     & 615.116      & 0.15        & 1284       \\
        &         & 0.001    & 10191.733     & 705.729      & 0.18        & 1108       & 10073.133     & 722.789      & 0.02        & 1166       \\
        & $n^2$   & 0.1      & 9489.7        & 619.521      & 0.3         & 1162       & 9463.933      & 803.496      & 0.47        & 1153       \\
        &         & 0.001    & 10023.833     & 480.342      & 0.84        & 1193       & 9431.233      & 810.004      & 0.01        & 1253       \\ \midrule
ca-AstroPh & $n^2/2$ & 0.1      & 15720.733     & 1155.778     & 0.25        & 1562       & 15162.233     & 1375.531     & 0.72        & 1476       \\
        &         & 0.001    & 16019.066     & 998          & 0.14        & 1438       & 15834.766     & 775.831      & 0.57        & 1586       \\
        & $n^2$   & 0.1      & 14374.766     & 1239.76      & 0.09        & 1292       & 13635.633     & 1485.423     & 0.01        & 1155       \\
        &         & 0.001    & 15083.133     & 956.011      & 0.88        & 1143       & 14354.866     & 1515.53      & 0.91        & 1106       \\ \bottomrule
\end{tabular}
}
\end{table}

Figures~\ref{fig:sp_vs_sg_l_uw} and \ref{fig:sp_vs_sg_s_uw} illustrate how the evaluated sampling-based weights evolve over increasing iterations. 
These figures also include the surrogate weights calculated for the generated solutions, derived through various approaches.
Notably, the figures demonstrate how the window in the \asg adjusts its size initially and then progresses alongside the weight scale.
A notable finding is that the weight evaluated using the surrogate method based on one-sided Chebyshev's inequality generally appears larger than that determined by other evaluation techniques.
This tendency is especially pronounced when $\alpha$ is small, a trend that is consistent with observations made in previous settings.
The sampling weights, on the other hand, are found to be closely aligned with the surrogate weights calculated using the Chernoff bound. 
Consequently, as the number of iterations increases, the algorithm employing the surrogate based on one-sided Chebyshev's inequality is likely to reach the bound earlier than other methods.

\section{Conclusion}
\label{sec:conclu}

In this paper, our investigation focuses on the \asg utilizing a sampling evaluation approach for chance-constrained monotone submodular problems,
specifically those with IID weights and uniform weights of identical dispersion.
In both settings, we assign larger values to the expected weight and dispersion.
Our experimental findings reveal that the \asg outperforms other algorithms like the \gs and \sg in solving the maximum coverage problem, 
especially in cases with larger bounds. 
Visual analyses show that the \asg is adept at incorporating a higher number of individuals within its adaptive window.
Additionally, it is observed that the algorithm's performance with a sampling-based evaluation is comparable in quality to that achieved with surrogate evaluations.

For future work, there is room for improvement in the algorithm's functionality, particularly regarding the window's performance in later iterations. 
Exploring the algorithms under different generalized settings and with various problems presents an interesting and valuable direction for future research.

\begin{acks}
This work has been supported by the Australian Research Council (ARC) through grant FT200100536.
\end{acks}

\balance
\bibliographystyle{ACM-Reference-Format}
\bibliography{reff}


\newpage
\appendix

\end{document}